\documentclass[sigconf]{acmart}
\usepackage{balance}
\usepackage{graphicx}
\usepackage{epstopdf}
\usepackage{booktabs} % For formal tables
\usepackage{amsmath}
\usepackage{amsmath}
\usepackage{bm}
\usepackage{algorithm}
\usepackage{algorithmic}
\usepackage{amsmath}
\usepackage{multirow}
\usepackage{indentfirst}
\usepackage{subfigure}
\usepackage{epsfig}
\usepackage{epstopdf}
\usepackage{xspace}

\usepackage{subeqnarray}

\newtheorem{myDef}{Definition}

\newcommand{\paratitle}[1]{\vspace{1.5ex}\noindent\textbf{#1}}
\newcommand{\ie}{\emph{i.e.,}\xspace}
\newcommand{\aka}{\emph{a.k.a.,}\xspace}
\newcommand{\eg}{\emph{e.g.,}\xspace}

\newcommand{\ignore}[1]{}

\def\BibTeX{{\rm B\kern-.05em{\sc i\kern-.025em b}\kern-.08emT\kern-.1667em\lower.7ex\hbox{E}\kern-.125emX}}

\begin{document}

\copyrightyear{2019}
\acmYear{2019}
\setcopyright{acmcopyright}
\acmConference[KDD '19]{The 25th ACM SIGKDD Conference on Knowledge Discovery and Data Mining}{August 4--8, 2019}{Anchorage, AK, USA}
\acmBooktitle{The 25th ACM SIGKDD Conference on Knowledge Discovery and Data Mining (KDD '19), August 4--8, 2019, Anchorage, AK, USA}
\acmPrice{15.00}
\acmDOI{10.1145/3292500.3330824}
\acmISBN{978-1-4503-6201-6/19/08}

% The "title" command has an optional parameter, allowing the author to define a "short title" to be used in page headers.
\title{Empowering $A^{*}$ Search Algorithms with Neural Networks for Personalized Route Recommendation}

% The "author" command and its associated commands are used to define the authors and their affiliations.
% Of note is the shared affiliation of the first two authors, and the "authornote" and "authornotemark" commands
% used to denote shared contribution to the research.
\author{Jingyuan Wang$^\ast$, Ning Wu}
\authornote{Both authors contributed equally to this work.}
\affiliation{%
  \institution{Beijing Advanced Innovation Center for BDBC,
School of Computer Science and Engineering,
Beihang University, Beijing, China}
%  \city{Beijing}
%  \state{China}
}\email{{jywang,wuning}@buaa.edu.cn}

%\author{Ning Wu}
%\authornotemark[1]
%\affiliation{%
%  \institution{MOE Engineering Research Center of ACAT,
%  School of Computer Science and Engineering, Beihang University}
%  \city{Beijing}
%  \state{China}
%}\email{wuning@buaa.edu.cn}

\author{Wayne Xin Zhao}
\authornote{Corresponding author.}
\affiliation{%
  \institution{School of Information, Renmin University of China}
  \city{Beijing}
  \state{China}
  }
\email{batmanfly@gmail.com}

\author{Fanzhang Peng, Xin Lin}
\affiliation{%
  \institution{MOE Engineering Research Center of ACAT,
  School of Computer Science and Engineering, Beihang University}
  \city{Beijing}
  \state{China}
}\email{{pengfanzhang,sweeneylin}@buaa.edu.cn}

%\author{Xin Lin}
%\affiliation{%
%  \institution{School of Computer Science and Engineering, Beihang University}
%  \city{Beijing}
%  \state{China}
%}\email{sweeneylin@buaa.edu.cn}

%
% By default, the full list of authors will be used in the page headers. Often, this list is too long, and will overlap
% other information printed in the page headers. This command allows the author to define a more concise list
% of authors' names for this purpose.
\renewcommand{\shortauthors}{Trovato and Tobin, et al.}

%
% The abstract is a short summary of the work to be presented in the article.
\begin{abstract}
Personalized Route Recommendation~(PRR) aims to generate user-specific route suggestions in response to users' route queries. Early studies cast the PRR task as a pathfinding problem on graphs, and adopt adapted search algorithms by integrating heuristic strategies.
Although these methods are effective to some extent,
they require  setting the cost functions with heuristics. In addition, it is difficult to utilize useful context information in the search procedure.
To address these issues, we propose using neural networks to automatically learn the cost functions of a classic heuristic algorithm, namely $A^{*}$ algorithm, for the PRR task. Our model consists of two components. First, we employ  attention-based Recurrent Neural Networks (RNN) to model the cost from the source to the candidate location by incorporating useful context information. Instead of learning a single cost value,  the RNN component is able to learn a time-varying vectorized representation for the moving state of a user. Second,  we propose to use a value network for estimating the cost from a candidate location to the destination. For capturing structural characteristics, the value network is built on top of improved graph attention networks by incorporating the moving state of a user and other context information.   The two components are integrated in a principled way for deriving a more accurate cost of a candidate location.
Extensive experiment results on three real-world datasets have shown  the effectiveness and robustness of the proposed model.
\end{abstract}

%
% The code below is generated by the tool at http://dl.acm.org/ccs.cfm.
% Please copy and paste the code instead of the example below.
%


%
% Keywords. The author(s) should pick words that accurately describe the work being
% presented. Separate the keywords with commas.

\maketitle

\section{Introduction}

With the popularization of GPS-enabled mobile devices, a huge volume of trajectory data from users has become available in a variety of domains~\cite{wang2017no,wang2018cd,wang2019understanding}. \emph{Personalized Route Recommendation~(PRR)} is one of the core functions in many online location-based applications, \eg online map. Given the road network, PRR aims to generate user-specific route suggestions on instant queries about the path planing from a source to a destination~\cite{cui2018personalized,dai2015personalized}.
It is challenging to perform effective pathfinding in a large and complex road network.
For accurate route recommendation, it is necessary to consider rich context information, including personalized preference, spatial-temporal influence  and road network constraint.%~\cite{cui2018personalized, dai2015personalized, lim2015personalized, Hsieh2012Exploiting, chen2016learning}.
%In response to the varying user queries, it is challenging to perform effective pathfinding in a large and complex road network.

Early studies cast the route recommendation task as a pathfinding problem on graphs~\cite{Wei2012Constructing,Luo2013Finding}.
These methods mainly focus on how to extend existing search algorithms (\eg Dijkstra shortest path algorithms and $A^{*}$ search algorithm)  for the studied task. With suitable heuristics, they can substantially reduce the search space and obtain high-quality responses. The key of heuristic search algorithms is to develop an effective cost function. Most of previous studies heuristically set the cost function, making their applicability highly limited. In addition, it is difficult to utilize various kinds of context information in the search process. To construct more flexible approaches, many studies have utilized machine learning methods for solving the PRR task~\cite{Wu2016Probabilistic,Chen2011Discovering}.
%, including inverse reinforcement learning~\cite{Wu2016Probabilistic}, absorbing Markov chain~\cite{Chen2011Discovering} and Gibbs sampling~\cite{banerjee2014inferring}.
These methods are able to characterize the location dependencies or spatial-temporal information with principled models. While, most of them are shallow computational models, and may have difficulties in capturing complex trajectory patterns. With the revival of deep learning, it sheds light on the development of more effective PRR models using neural networks. Especially, sequential neural models, \ie Recurrent Neural Networks (RNN), have been widely used for modeling sequential trajectory data~\cite{Wu2017Modeling, Al2017STF, Yang2017}. However, to our knowledge, these models mainly focus on one-step or short-term location prediction, which may not be suitable for the PRR task.

Comparing the above approaches, we can see they have their own merits for the PRR task. On one hand, in terms of the problem setting, heuristic search algorithms are specially suitable for the PRR task, which can be considered as a pathfinding problem on graphs given the source and destination. They are able to generate high-quality approximate solutions using elaborate heuristics. On the other hand, as a newly emerging direction of machine learning, deep learning methods are effective to capture the complex data characteristics using learnable neural networks. They are able to learn effective mapping mechanisms from input to output or expressive feature representations from raw data in an automatic way.  For developing a more effective PRR method, is there a principled way to combine the merits of both kinds of approaches?

Inspired by recent progress of deep learning in strategy-based games (\eg Go and Atari)~\cite{silver2016mastering,mnih2015human} , we propose to improve search algorithms with neural networks for solving the PRR task. Especially, we adopt the $A^{*}$ algorithm~\cite{a_start} as the base search algorithm, since it has been widely used in pathfinding and graph traversal. Previous studies have also shown that $A^{*}$ algorithm is a promising approach to solving the route recommendation task~\cite{Wei2012Constructing, pallottino1998shortest, kanoulas2006finding}. The main idea of our solution is to automatically learn the cost functions in $A^{*}$ algorithms, which is the key of heuristic search algorithms. For this purpose, there are three important issues to consider. First, we need to define a suitable form for the cost in the PRR task. Different from traditional graph search problems, a simple heuristic form can not directly optimize the goal of our task~\cite{kanoulas2006finding, Wei2012Constructing}, \eg the route based on the shortest distance may not meet the personalized needs of  a specific user. Second, we need to design effective models for implementing cost functions with different purposes, and unify different cost functions for deriving the final cost. The entire cost function $f(\cdot)$ of $A^{*}$ can be decomposed into two parts, \ie  $f(\cdot) = g(\cdot)+h(\cdot)$. The two parts compute the \emph{observable cost} from the source node to the evaluation node and the \emph{estimated cost} from the evaluation node to the destination node respectively. Intuitively, the two parts require different modeling methods, and need to jointly work to compute the entire cost. Third, we need to utilize rich context or constraint information for improving the task performance. For example, spatial-temporal influence and road networks are important to consider in modeling trajectory patterns, and should be utilized to develop the cost functions.

To address these difficulties, we propose a novel neuralized $A^{*}$ search algorithm for solving the PRR task. To define a suitable form for the search cost, we formulate the PRR task as a conditional probability ranking problem, and compute the cost by summing the negative log of conditional probabilities for each trajectory point in a candidate trajectory.
%In this way, we can sum the negative log of conditional probabilities from each trajectory point for computing the cost of a candidate trajectory.
 We use this form of cost to instruct the learning of the two cost functions in $A^{*}$ algorithm, namely  $g(\cdot)$ and $h(\cdot)$. For implementing $g(\cdot)$, we propose to use attention-based RNNs to model the trajectory from the source location to the candidate location.  We incorporate useful context information to better capture sequential trajectory behaviors, including spatial-temporal information, personalized preference and road network constraint. Instead of simply computing a single cost, our model also learns a time-varying vectorized representation for the moving state of a user. For learning $h(\cdot)$, we propose to use a value network for estimating the cost for unobserved part of a trajectory. In order to capture the complex characteristics of road networks, we build the value network on top of improved graph attention networks by incorporating useful context information. In these two different components, we share the same embeddings for locations, users and time. The learned moving state in the RNN component will be subsequently fed into the value network.
%Our RNN component is also used to compute the immediate reward for the value network.
The two components are integrated in a joint model for computing a more accurate cost of a candidate location. Since the estimated cost is associated with a multi-step decision process, we further propose to use the \emph{Temporal Difference} method from Reinforcement Learning for model learning.

%Our contribution can be summarized as follows.

To the best of our knowledge, we are the first to use neural networks for improving $A^{*}$ algorithm in the PRR task. Our approach is able to automatically learn the cost functions without handcrafting heuristics. It is able to effectively utilize context information and characterize complex trajectory characteristics, which elegantly combines the merits of $A^{*}$ search algorithms and deep learning.
The two components are integrated in a joint model for deriving the evaluation cost. Extensive results on the three datasets have shown the effectiveness and robustness of the proposed model.

\section{RELATED WORK}

Our work is related to the following research directions.

\paratitle{Route Recommendation.}  With the availability of user-generated trajectory information,  route recommendation has received much attention from the research community~\cite{kanoulas2006finding,cui2018personalized, dai2015personalized}, which aims to generate reachable paths between the source and destination locations. The task can be defined as either \emph{personalized}~\cite{cui2018personalized, dai2015personalized} or \emph{non-personalized}~\cite{kanoulas2006finding, yuan2010t, Luo2013Finding, Chen2011Discovering}, and constructed based on different types of trajectory data, \eg GPS data~\cite{yuan2010t} or POI check-in data~\cite{shafique2016recommending,chen2016learning}. In the literature, various methods have been developed for route recommendation, including graph search algorithms~\cite{Chen2011Discovering,Wei2012Constructing, Liu2011Route}, time-sensitive algorithms~\cite{Luo2013Finding}, $A^{*}$ search algorithm~\cite{kanoulas2006finding}, probabilistic POI transition/ranking models~\cite{chen2016learning} and diver-direction based methods~\cite{yuan2010t}.
Overall, most of the studies focus on using search based algorithms or probabilistic models by considering additional constraints, \eg road networks or time.
Our work is built on top of search based solutions, and the novelty lies in the automatic  learning of the cost functions using neural networks.
Our model is  flexible to incorporate rich context or constraint information.

\paratitle{Deep Learning for Trajectory Data Mining.}
Recent years have witnessed the success of deep learning in modeling complex data relations or characteristics.
In specific,  Recurrent Neural Network (RNN)  together with its variant Long Short-Term Memory (LSTM) and Gated Recurrent Unit (GRU) have been widely used for modeling sequential trajectory data.
Typical works include hierarchical RNN~\cite{zheng2017generating}, RNN with  road network constraints~\cite{Wu2017Modeling}, and multi-modal embedding RNN~\cite{feng2018deepmove}, spatial-temporal RNN~\cite{Liu2016Predicting} and space time feature-based RNN~\cite{Al2017STF}. These studies mainly focus on short-term
trajectory behaviors, \eg
one-step location recommendation~\cite{Liu2016Predicting}, which are not suitable for  solving the current task.

%long-term trajectory generation~\cite{zheng2017generating} or one-step location recommendation~\cite{Liu2016Predicting}, which are not suitable for  solving the current task.

\ignore{ Zheng et al. propose a hierarchical RNN to generate Long-term trajectories~\cite{zheng2017generating}. Wu et al. introduce a novel RNN model constrained by the road network to model trajectory~\cite{Wu2017Modeling}.
 Feng et al. design a multi-modal embedding recurrent neural network with historical attention to capture the complicated sequential transitions~\cite{feng2018deepmove}.
 Chang et al. employ the RNN and GRU models to capture the sequential relatedness in mobile trajectories at different levels~\cite{Yang2017}.
 Liu et al. extend RNN and propose a novel method called Spatial Temporal Recurrent Neural Networks to predict the next location of a trajectory~\cite{Liu2016Predicting}.
 Al-Molegi et al. propose a novel model called Space Time Features-based Recurrent Neural Network (STF-RNN) for predicting people next movement based on mobility patterns obtained from GPS devices logs~\cite{Al2017STF}.}

\paratitle{Machine Learning for Heuristic Search.} These studies in this direction aim to automatically improve or optimize the search algorithms with machine learning methods.
Early works include the use of machine learning in creating effective, likely-admissible or improved heuristics~\cite{lelis2011predicting,ernandes2004likely,samadi2008learning}.
More recently, deep learning has significantly pushed forward the research of this line.
The main idea is to leverage the powerful modeling capacity  of neural networks for improving the  tasks that require complicated solving strategies, including the Go game~\cite{silver2016mastering} and Atari games~\cite{mnih2015human}.
Our work is highly inspired by these pioneering works, but have a quite different focus on the studied task, \ie personalized route recommendation.
Our task itself involves specific research challenges that make the reuse of previous works impossible.

\section{PRELIMINARIES}

In our task, we assume road network information is available for the pathfinding task, which is the foundation of the traffic communication for users.

\begin{myDef}
\textbf{Road Network}. A road network is a directed graph $\mathcal{G}=(\mathcal{L},\mathcal{E})$, where $\mathcal{L}$ is a vertex set of locations and $\mathcal{E} \subset \mathcal{L} \times \mathcal{L}$ is an edge set of road segments. A vertex $l_i \in \mathcal{L}$ (\ie a location) represents a road junction or a road end. An edge $e_{l_i, l_j}=\langle l_i, l_j \rangle \in \mathcal{E}$ represents a directed road segment from vertex $l_i$ to vertex $l_j$.
%We further use $\mathcal{L}_{l_i}$ to denote all the connected vertices for $l_i$ by one-step road segment.
\end{myDef}

\begin{myDef}
\textbf{Route}. A route  (\aka a path) $p$ is an ordered sequence of locations 
connecting the source location $l_s$ with the destination location $l_d$ with $m$ intermediate locations, 
\ie $p:  l_s\rightarrow l_1 \rightarrow ...\rightarrow l_m \rightarrow l_d$, where
each   pair of consecutive locations $\langle l_{i}, l_{i+1} \rangle$ corresponds to a road segment $e_{l_i,l_{i+1}}$ in the road network.
%connected by road segments starting from $l_s$ and ending at $l_d$, \ie $p:  l_s\rightarrow l_1 \rightarrow ...\rightarrow l_n \rightarrow l_d$, where each  consecutive pair $\langle l_{i}, l_{i+1} \rangle$ corresponds to a road segment $e_{l_i,l_{i+1}}$.
%, and $l_s$ and $l_e$ are the start and ending locations for $p$. %In other words, $l_{i}$ and $l_{i+1}$ are the endpoints of road segment $r_k$.
\end{myDef}

The moving trajectory of a user on the road network can be recorded using GPS-enabled devices. Due to instrumental inaccuracies, the sampled trajectory points may not be well aligned with the locations in $\mathcal{L}$. Following~\cite{yang2018fast}, we can preform the procedure of \emph{map matching} for aligning trajectory points with locations in $\mathcal{L}$.

\ignore{Hence, we preform the proprocessing procedure of \emph{map matching} for aligning trajectory points with locations in the road network.
Following~\cite{}, we map a trajectory point to one of the two endpoints of a road segment according to the geographical distance.
After that, each sampled trajectory point can be mapped into an exact location in $\mathcal{L}$.
Next, we are ready to characterize a trajectory sequence generated by a user.
}
%Given the road network, we can record the moving behaviors of a user as a trajectory sequence consisting of time-ordered trajectory points using GPS-enabled devices.

%\begin{myDef}
%\textbf{Map Matching}. Map matching~\cite{} is a standard preprocessing procedure for mapping trajectory points to road segments. Following~\cite{}, we map a trajectory point to an endpoint
%of a road segment according to the geographical distance.
%\end{myDef}

%After map matching, we are ready to characterize a trajectory sequence generated by a user.

\begin{myDef}
\textbf{Trajectory}. A trajectory  $t$ is a time-ordered sequence of $m$ locations (after map matching) generated by a user, \ie $t: \langle l_1, b_1 \rangle\rightarrow \langle l_2, b_2 \rangle\rightarrow ...\rightarrow \langle l_m, b_m\rangle$, where $b_i$ is the visit timestamp for location $l_i$.
\end{myDef}

%A location $l$ is associated with a triple $(x,y,id)$, where $l.x$ is the longitude, $l.y$ is the latitude, and $l.id$ is an integer identifier corresponding to a .

%Above, a route is modeled as a location sequence, and it can be also modeled as a sequence of road segments, where consecutive road segments share the same endpoint.

A trajectory is a user-generated location sequence with timestamps, while a route is pre-determined by the road network.
For a route, the start and end points are important to consider, which correspond to the source and destination of a travel. In the task of PRR, a user can issue
\emph{travel queries}.

\begin{myDef}
\textbf{Query}. A query $q$ is a triple $\langle l_s, l_d, b\rangle$ consisting of source location $l_{s}$, destination location $l_{d}$ and departure time $b$.
\end{myDef}

With the above definitions, we now define the studied task.

\begin{myDef}
\textbf{Personalized Route Recommendation (PRR)}. Given a dataset $\mathcal{D}$ consisting of historical trajectories, for a query $q: \langle l_s, l_d, b\rangle$ from user $u \in \mathcal{U}$, we would like to infer the most possible route $p^{*}$ from $l_{s}$ to $l_{d}$ made by user $u$, formally defined as solving the optimal path with the highest conditional probability:
\begin{equation}\label{eq-prrtask}
p^{*} = \arg\max_{p} \text{Pr}(p | q, u, \mathcal{D}).
\end{equation}
\end{myDef}

%\paratitle{Problem statement.} Suppose we have a dataset $\mathcal{D}$ consisting of trajectories, given a query $q$, we would like to infer the most satisfied route $\hat{t}$ corresponding the query $q$.

\ignore{\begin{myDef}
\textbf{Location}. A location $l$ is stored by a tuple $(x,y,id)$, where $x$ is the longitude, $y$ is the latitude, and $id$ is the id of an edge on road network $G$.
\end{myDef}
\begin{myDef}
\textbf{Record}. A record $e_i$ is from an user $u$ and recorded by a tuple $(l,s)$, where $l$ is the location and $s$ is the timestamp of the record.
\end{myDef}
\begin{myDef}
\textbf{Trajectory}. A trajectory $T$ is a time-ordered sequence of $n$ locations generated by an user $u$, i.e., $T=(t,u),t=e_1\rightarrow e_2 \rightarrow ...\rightarrow e_n$. Given a trajectory $T$, we have $e_{i+1}.s - e_{i}.s=\epsilon$ and $\epsilon$ is a constant sampling interval.
\end{myDef}
\begin{myDef}
\textbf{Route}. A route $R$ is a time-ordered sequence of $n$ locations generated by an user $u$, i.e., $R=(r,u),r=l_1\rightarrow l_2 \rightarrow ...\rightarrow l_n$
\end{myDef}
\begin{myDef}
\textbf{Query}. A query $q$ is a tuple $(o, d, s, u)$ that consists of origin $o$, destination $d$, departure time $s$ and user $u$.
\end{myDef}
\begin{myDef}
\textbf{Most Satisfied Route}. A route $\hat{R}$ inferred by query $q$ that minimize the distance between $\hat{R}$ and real route $R$ from $u$.
\end{myDef}
}

The PRR task is formulated as a conditional ranking problem.
For solving this task, we first present a traditional $A^{*}$-based algorithm in Section 4, and then present our proposed approach in Section 5. 

%\begin{problem}

  \section{A Heuristic $A^{*}$ Solution for PRR}

  The task of PRR can be framed as a graph-based search problem.
  In this setting, we view the road network as a graph, and study how to find possible route(s)
  that start from the source node and end at the destination node.
  
  \paratitle{Review of $A^{*}$ Algorithm.} In the literature~\cite{a_start}, $A^{*}$ search algorithm is widely used in pathfinding and graph traversal due to its performance and accuracy. Starting from a source node of a graph, it aims to find a path to the given destination node resulting in the smallest \emph{cost}.
  It maintains a tree of paths originating at the source node and extending those paths one edge at a time until its termination criterion is satisfied.
  At each extension, $A^{*}$  evaluates a candidate node $n$ based on a \emph{cost function} $f(n)$
  \begin{equation}
  f(n) = g(n) + h(n),
  \end{equation}
  where $g(n)$ is the cost of the  path from the source to $n$ (we call it \emph{observable cost} since the path is observable), and  $h(n)$ is an estimate of the cost required to extend the future path to the goal (we call it \emph{estimated cost} since the actual optimal path is unknown).
  %For convenience, we call the value that $g(n)$ computes \emph{observable cost}, since it is observable for the path from source
  %These two parts essentially compute the  \emph{observable cost} and  \emph{estimated cost} respectively.  %In $A^{*}$ algorithm, the cost function $h(\cdot)$ is very important, which directly determine the final performance. $h(\cdot)$ computes the estimated cost from a candidate node to the destination.
  The key part of $A^{*}$  is the setting of the heuristic function $h(\cdot)$, which has an important impact on the final performance.
  %A heuristic function $h(\cdot)$ is called \emph{admissible} if it never overestimates the actual minimal cost of reaching the goal.
  %For tree search problems, if an admissible heuristic is used, the $A^{*}$ search algorithm will never return a suboptimal goal node~\cite{hart1968formal}.
  
  \paratitle{A Simple $A^{*}$-based Approach for PRR.}  Considering our task, the goal is to maximize the conditional probability of $\text{Pr}(p | q, u, \mathcal{D})$. We can equally minimize its negative log: $-\log \text{Pr}(p | q, u, \mathcal{D})$. %Furthermore, let $p^{*}: l^{*}_1 \rightarrow l^{*}_2 \cdots \rightarrow l^{*}_m $ denote the optimal route.
  Given a possible path $p: l_{s} \rightarrow l_1 \rightarrow l_2 \cdots \rightarrow l_m \rightarrow l_d$, consisting of $m$ intermediate locations,
  we can factorize the path to compute its cost according to the chain rule in probability in the form of
  \begin{eqnarray}\label{eq-optimalcost}\small
  %&&-\log \text{Pr}(p | q, u, \mathcal{D}),\\\nonumber
  -\log \text{Pr}\left(p | q, u, \mathcal{D}\right)=-\sum_{i=0}^{m}\log  \text{Pr}\left(l_{i+1} | l_s\rightarrow l_{i}, q, u\right),
  \end{eqnarray}
  where  $l_0=l_s$ and $l_{m+1}=l_d$, and $ \mathcal{D}$ is dropped for simplifying notations. This  formula  motivates us to set the cost functions of $A^{*}$ algorithm in a similar form.
  Assume a partial route has been generated, \ie $p: l_{s} \rightarrow l_1 \cdots \rightarrow l_{i-1}$, we can compute the observable cost of a candidate $l_i$  for extension as
  \begin{eqnarray}\label{eq-gcost}\small
  g\left(l_s \rightarrow l_i\right)=-\sum_{k=1}^{i-1}\log  \text{Pr}\left(l_{k+1} | l_s\rightarrow l_{k}, q, u\right).
  \end{eqnarray}
  To compute the conditional transition probabilities, the first-order Markov assumption is usually adopted,
  so we  have $\text{Pr}(l_{k+1} | l_s\rightarrow l_{k}, q, u) = \text{Pr}(l_{k+1} | l_{k}, q, u)$.
  Following~\cite{Wei2012Constructing, Chen2011Discovering}, we can further use user-specific or overall transition statistics to estimate
  the probabilities (with smoothing).
  While, to compute the cost of $h(l_i \rightarrow l_d)$ is more difficult, since the optimal sub-route from $l_i$ to $l_d$ is unknown. We cannot directly apply the similar method in Eq.~\eqref{eq-gcost} for $h(\cdot)$. In practice, we can use different heuristics to set $h(\cdot)$, including
  the shortest spatial distance~\cite{nachtigall1995time, a_start} and the historical likelihood~\cite{Wei2012Constructing}.

  \ignore{
  The satisfaction function of the inferred route can be denoted as $S(\hat{r},q)$, it can be equally represented as
  \begin{equation}
  \begin{aligned}
  S(\hat{r},q)=S_1(l_1, q)  + \sum_{i=2}^k S_i(r_{i-1}, l_i, q)
  \end{aligned}
  \end{equation}
  where $r_i$ is the generated route $\{l_1, l_2...l_{i}\}$. In previous work\cite{}, to simplify the complexity of $S_i(l_i, r_i,q)$, it is often replaced by $S_i(l_i, l_{i-1})$, and $S_i(l_i, l_{i-1})$ can be calculated by the formula
  $$
  S_i(l_i, l_{i-1})=\frac{N_{l_i \rightarrow l_{i-1}}}{N_{l_{i-1}}}
  $$
  where $N_{l_i \rightarrow l_{i-1}}$ is the number of trajectories passing $l_i \rightarrow l_{i-1}$, while $N_{l_{i-1}}$ is the number of trajectories passing $l_{i-1}$.
  To search the most satisfied route for user by $A^{*}$ algorithm, $Sa(l_i, l_{i-1})$ can be consider as the distance between $l_i$ and $l_{i-1}$. Furthermore, since the distance has been replaced by satisfaction, the typical heuristics value used to guide the $A^{*}$ search procedure such as Manhattan distance also need to be substituted by other hand-designed heuristic value. Hence, the $f$ score can be calculated by:
  $$
  f = S(\hat{r}_i) + H(\hat{r}, l_n)
  $$
  }
  
  \paratitle{Analysis}. For our task, the $A^{*}$-based approach is more appealing than a greedy best-first search algorithm. By decomposing the entire cost into two parts, it leaves room on the elaborated setting of $g(\cdot)$ and $h(\cdot)$ for different tasks. Although
  it has been shown that $A^{*}$-like algorithms perform well in the task of route recommendation~\cite{Wei2012Constructing, nachtigall1995time, kanoulas2006finding}, we see three weak points for improvement. First,  $A^{*}$ algorithm is a general framework in which cost functions have to be heuristically set. It is difficult to incorporate varying context information, \eg personalized preference and spatial-temporal influence. Second, the cost function usually relies on the heuristic computation or estimation, which is easy to suffer from  data sparsity. For example, the estimation of transition probabilities in Eq.~\eqref{eq-gcost} may not be accurate when the historical transitions between two locations are sparse. In this case, even the computation of observable cost $g(\cdot)$ is likely to be problematic. Third, the PRR task is challenging, and a simple heuristic search strategy may not be capable of performing effective pathfinding in practice, \eg the route that has the shortest spatial distance may not be the final choice of a user~\cite{Chen2011Discovering, Luo2013Finding}. With these considerations, we next present our solution for addressing the above difficulties of $A^{*}$ in PRR.
  %However, that is not a good way to implement personalized recommendation and model spatial-temporal information. In fact, learning dynamic satisfaction and heuristics value
  %is a better choice and deep neural network can be used to model complex spatial-temporal and user pattern. Hence, we utilize a satisfaction network and a heuristics network to approximate two value.
  
  \section{The NASR Model}
  
  %\section{Empowering the A$^{*}$ Search Algorithm with Neural Networks}
  
  In the section, we present the proposed \emph{Neuralized A-Star based personalized route Recommendation (NASR)} model.

  %a novel neuralized $A^{*}$ search algorithm, which combines both merits of  $A^{*}$ search and neural networks, for personalized route recommendation.  We name it as \emph{Neuralized A-Star based personalized route Recommendation (NASR)}.
  %We first present the overall model overview, and then describe each part in detail.
  
  %heuristics search algorithm assisted by deep learning for route recommendation, called HDRec.
  %First, we state the framework of dynamical A* search, and talk about the combination between traditional A* search and deep learning. Second, we introduce the multi-context heuristics neural network consists of multi-context embedding module, feature extraction module and output module. Finally, we introduce the neighbour-constrained margin loss and deep Q learning to learn heuristics $g$ and $h$ in A* search respectively.
  
  \begin{figure}[t]
   \includegraphics[width=0.85\linewidth]{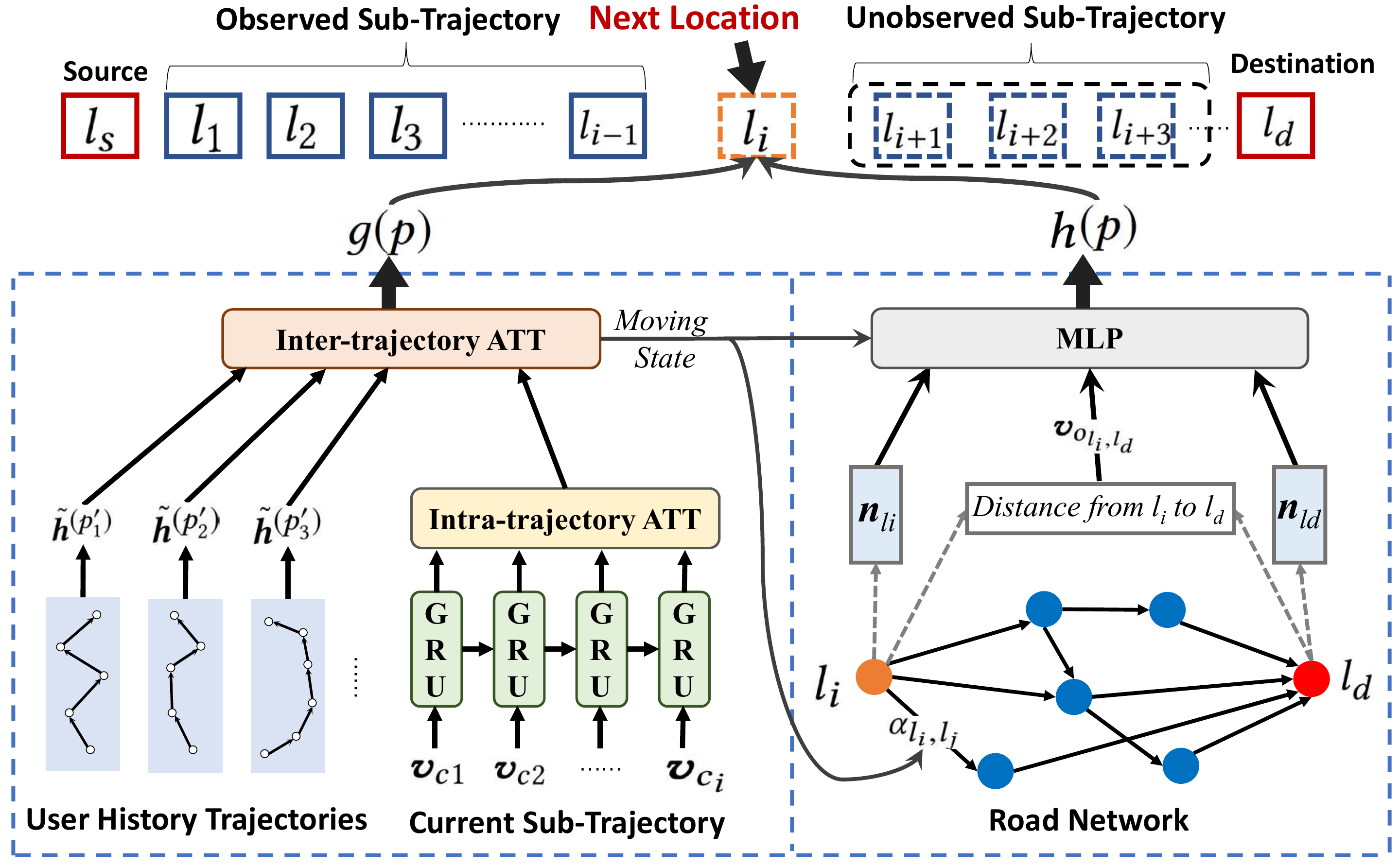}\vspace{-.3cm}
   \caption{The overall architecture of the NASR model. $g(\cdot)$ learns the cost  from the source to a candidate location, called \emph{observable cost}; $h(\cdot)$ predicts the estimated cost from a candidate location to the destination, called \emph{estimated cost}.
   }
   \label{fig-framework}\vspace{-.3cm}
  \end{figure}

  \subsection{Model Overview}
  Our model is developed based on the general $A^{*}$ algorithm framework.
  For node evaluation, we decompose the entire cost function $f(\cdot)$  into two parts, namely \emph{observable cost} and \emph{estimated cost}, which correspond to the cost functions $g(\cdot)$ and  $h(\cdot)$.
  Traditionally, both $g(\cdot)$ and  $h(\cdot)$ are heuristically computed or set. While,
  our idea is to automatically learn the two functions with neural networks instead of using heuristics.
  Specially, we use Recurrent Neural Networks (RNN) to implement $g(\cdot)$
  and another value network
   to implement $h(\cdot)$.
  %which directly learns the cost  from the source to a candidate location. Meanwhile, we use another value network
   %to implement $h(\cdot)$ for estimating the cost from a candidate location to the destination.
  In our neural network for function $g(\cdot)$, we not only compute a single  cost value, but also learn
  a time-varying  \emph{moving state} for a specific user.
  The moving state encodes necessary trajectory information of a user till the evaluation time, which will be fed into the computation of $h(\cdot)$.
  Once the two networks are learned, we can compute the cost of a candidate location for path extension.   We present the overall architecture for the proposed model in Fig.~\ref{fig-framework}.

  \subsection{Modeling the Observable Cost with RNN} %Recurrent Neural Networks}
  This part studies the learning of function $g(\cdot)$ for observable cost.
  Given an observed sub-route $l_s \rightarrow l_1 \rightarrow l_2 \cdots \rightarrow l_i$, as shown in Eq.~\eqref{eq-gcost}, the problem becomes how to effectively learn the conditional transition probabilities $\text{Pr}(l_{k+1} | l_s \rightarrow l_{k}, q, u)$. Simple frequency-based estimation method will suffer from data sparsity in large search space even with first-order Markov assumption. In this case, the computed observable cost will not be reliable to be used.
  In addition, PRR is a user-centric task, and a single cost value may not
  be enough to describe what has been observed.
  Instead, the \emph{moving state} of a user and useful context information should be considered.
  %In addition, our task aims to learn personalized trajectory patterns for a specific user, and many useful context information should be considered for cost computation.
  To address these difficulties, we propose to use Recurrent Neural Networks to implement the cost function $g(\cdot)$.
  %model observed sub-routes and automatically learn transition probabilities.
  
  %Unlike traditional pathfinding tasks, \eg the shortest path search, we do not necessarily recommend the shortest route in spatial space. Instead, we try to learn personalized trajectory patterns for a specific user. Hence, a simple frequency-based method in Eq.~\ref{eq-factorizedprob} for the observed travel is not the best strategy in our task.

  %We propose to implement $g(n)$ with the measuring network.
  %Another novelty from our model is that we can learn the travel state of a user and then leverage such state information for estimating future cost.
  %With the measuring network, we can effectively incorporate various of context information for user modeling, including historical trajectories, spatial-temporal information, and even user profiles.
  
  %The neural network that is used to estimate heuristics consists of three modules. The first module is the multi-source information embedding module. By the module, location, distance, time and user information are extracted primarily, then they will be feed into the feature extraction module. In the high level feature extraction procedure, GRU is applied to model the complex sequential transition and model history trajectory. Finally, satisfaction and heuristics are trained by neighbour constrained loss and reinforcement learning.
  
  \subsubsection{Embedding Rich Context Information}
  As the prerequisite module, we embed rich context information into dense vectors, which will be subsequently used by other components.
  First, we set up an embedding vector $\bm{v}_u \in \mathbb{R}^{K_U}$ for user $u$, encoding necessary personalized user information. Then, for each location $l \in \mathcal{L}$, we set up a corresponding embedding vector $\bm{v}_l \in \mathbb{R}^{K_L}$.
  For trajectory behaviors, temporal information is also important to consider. Following~\cite{feng2018deepmove}, for each visit timestamp $b$, we use two embedding vectors $\bm{v}_{di(b)} \in \mathbb{R}^{K_D}$ and $\bm{v}_{hi(b)} \in \mathbb{R}^{K_H}$, where $di(b)$ and $hi(b)$ are functions transforming $b$ into corresponding weekday index ($1$ to $7$) and hour index ($1$ to $24$) respectively.
  %These embedding vectors have different dimension and are used in all components in our model.
  At $i$-th time step, we concatenate the above embedding vectors  into a single embedding vector, and form an enhanced representation of context $x_i$ as
  \begin{equation}\label{eq-vci}
    \bm{v}_{x_i} =
    \bm{v}_u  \Vert  \bm{v}_{l_i} \Vert \bm{v}_{di(b_i)} \Vert \bm{v}_{hi(b_i)},
  \end{equation}
  where ``$\Vert$" is the concatenation operation. We can see that $\bm{v}_{x_i}$ contains information for user preference, current location and time.
  
  %In the module, we will embed temporal, location and user information into dense vector, these representation will be shared in feature extraction module. For temporal information, we divide 24 hour into 96 time slices and denote the id of time slice as $t_i^m$, besides, weekday and weekend id is denoted as $t_i^w$. Their corresponding representation $\bm{t_i^m}$ and $\bm{t_i^w}$ will be concated into a new vector $\bm{t_i}$. There are some user information including id, age, job etc. These information are denoted as $u_p$ and it will be embedded as $\bm{u_p}$. Similarily, location $l_i$ also will be embedded as $\bm{l_i}$. At every timestamp, time embedding, location embedding and user embedding will be concated into a new $\bm{l_i}$ as the input of sequential feature extraction module and attention module.
  
  \subsubsection{Encoding the Observed Sub-Trajectory with RNN}
  For the PRR task, it is important to model the trajectory characteristics of users' moving behaviors, which can be considered as a sequential process. We utilize RNNs to model such sequential behaviors. Given an observed sub-trajectory $p: l_s \rightarrow l_1 \cdots \rightarrow l_i$ generated by $u$, we employ the widely used GRU network~\cite{gru} to encode it into a vector
  \begin{equation}%\small
  \bm{h}_{i}^{(p)} = \text{GRU}(\bm{v}_{x_i}, \bm{h}_{i-1}^{(p)}),  \label{4}
  \end{equation}
  where $\bm{h}_{i}^{(p)}  \in \mathbb{R}^{K_R}$ is the hidden vector produced by the GRU network and $\bm{v}_{x_i}$ is the context vector defined in Eq.~\eqref{eq-vci}. The vector $\bm{h}_{i}^{(p)}$ encodes the \emph{moving state} of a user at the $i$-th time step.
  Note that we use the superscript to index the trajectory and the subscript to index locations. Unlike traditional $A^{*}$ search algorithm, here, we learn an informative state representation of a user at each step, providing more information than a single cost value.
  
  \subsubsection{Enhanced Moving States with Attention Mechanism}
  An observed sub-trajectory can be short and noisy. We further propose to use
  two types of attention to
   improve the learning of moving state  by leveraging data dependence.
  %We consider two types of attention for capturing trajectory characteristics.
  %In history attention module, we still introduce GRU as base feature extraction module. For every user, every location in his history trajectory will be embedded into a vector by GRU, which is denoted as $\bm{h}^u_{m,n}$, where $n$ is the $n$-th trajectory and $m$ is the $m$-th location in trajectory.

  \paratitle{Intra-Trajectory Attention}. We first apply the method~\cite{bahdanau2015neural} to compute the attention between locations in the same trajectory as
  \begin{align}
  \bm{\tilde{h}}_i^{(p)} =\sum_{k=1}^{i}\text{att}\left(\bm{h}_i^{(p)},\bm{h}_k^{(p)}\right)\cdot \bm{h}_k^{(p)},
  \end{align}
  where $\bm{\tilde{h}}_i^{(p)} $ denotes the improved state representation with intra-trajectory attention and $\text{att}(\cdot, \cdot)$ is an attention function as
  \begin{eqnarray}\label{eq-att}\small
  \text{att}\left(\bm{h}_i^{(p)},\bm{h}_k^{(p)}\right)&=& \frac{\alpha_{i,k}}{\sum_{k'=1}^{i}\exp(\alpha_{i,k'})},  \\\nonumber
  \alpha_{i,k}&=& {\bm{w}}_1^{\top} \cdot \tanh\left(\bm{W}_1 \cdot \bm{h}_{k}^{(p)} + \bm{W}_2 \cdot \bm{h}_i^{(p)}\right),\nonumber
  \end{eqnarray}
  where $\bm{w}_1$, $\bm{W}_1$ and $\bm{W}_2$ are the parameter vector or matrices to learn.
  With intra-trajectory attention, we can discover more important characteristics by considering the entire trajectory. After intra-trajectory attention, we use the state representation of the last location for encoding the entire sub-trajectory, \ie $\bm{\tilde{h}}^{(p)}=\bm{\tilde{h}}_i^{(p)}$.
  %To measure the satisfaction of next location, we can calculate the scalar product $\bm{l_{i-1}}^T \bm{l_{i}}$ between current location embedding $\bm{l_{i-1}}$ and next location embedding $\bm{l_i}$, which represent the semantic distance between current location and next location. User information and time information are both introduced by the vector that consists of semantic distance, user information and time information.
  
  %\begin{equation}
  %\begin{aligned}
  %\bm{sa_{d}} =  [\bm{l_{i-1}}^T \bm{l_{i}};  dist(l_{i-1}, l_{i})]\bm{w_{sd}} + b_{sd}
  %\end{aligned}
  %\end{equation}
  %\bm{u_p}; \bm{t_i};
  %Similarily, to estimate the heuristics value, we need to calculate the semantic distance between current location and destination, the semantic distance can be formulated as:
  
  %\bm{u_p}; \bm{t_i};
  %\begin{equation}
  %\begin{aligned}
  %\bm{he_{d}} =  [\bm{l_{i}}^T \bm{l_T};  dist(l_{i}, l_n)]\bm{w_{hd}} + b_{hd}
  %\end{aligned}
  %\end{equation}
  
  \paratitle{Inter-Trajectory Attention}. The information from a single trajectory is usually limited.
  In order to capture overall moving patterns for a specific user,
   we further  consider incorporating historical  trajectories generated by the user. Given the current trajectory $p$, we attend it to each of the other historical trajectories as
  \begin{equation}\label{eq-hstate}\small
  \bm{{h}}^{(p)} =\sum_{p' \in \mathcal{P}^u } \text{att}\left(\bm{\tilde{h}}^{(p)}, \bm{\tilde{h}}^{(p')}\right)\cdot \bm{\tilde{h}}^{(p')},
  \end{equation}
  where $\mathcal{P}^u$ denotes the set of historical trajectories generated by $u$ and $\text{att}(\cdot, \cdot)$ is the attention function which is similar to Eq.~\eqref{eq-att} but with different learnable parameters. %The inter-trajectory attention is useful to
  
  %, we can collectively learn effective characteristics using all the trajectory data from a user.

  %$$
  %L_{nei}=\sum_{i}^{N} \max(0, 1 -  g(l_i, r_{i-1}, q) + \max_{l^{'} \in neighs-l_i} g(l^{'}, r_{i-1}, q))
  %$$
  
  \subsubsection{Observable Cost Computation with the  Road Network Constraints}
  Once we have learned the hidden state for the current timestamp, we are able to compute the probability of the next location using a softmax function with road network constraint as
  \begin{equation}\label{eq-prli}\small
  \text{Pr}(l_i | l_{s} \rightarrow l_{i-1}, q, u)=\frac{\exp\left(z(p^{l_s \rightarrow l_{i}})\right)}{\sum_{l' \in \mathcal{L}_{l_{i-1}}}\exp\left(z(p^{l_s \rightarrow l'})\right)},
  \end{equation}
  where $z(p) = \bm{w}_2^\top\cdot\bm{{h}}^{(p)}$ is a linear transformation function taking as input the hidden state learned for a trajectory $p$ in Eq.~\eqref{eq-hstate}, and $\bm{w}_2$ is parameter vector to learn.
  %The $\mathcal{L}_{l_{i-1}}$ is the set of the neighboring locations with $l_i$ in the road networks.
  Here, we compute the probability of a candidate location $l_i$ by normalizing over all the neighboring locations of $l_{i-1}$ in the road network. After defining $\text{Pr}(l_i| {l_s \rightarrow l_{i-1}}, q, u)$ in Eq.~\eqref{eq-prli}, we sum the negative log probability of each location in a trajectory $l_s \rightarrow l_{i}$ as the value of $g(\cdot)$
  \begin{equation}\label{eq-gdef}\small
  g(l_s \rightarrow l_i) = -\sum_{j=2}^{i}\log \text{Pr}\left(l_{j}|{l_s \rightarrow l_{j-1}}, q, u\right).
  \end{equation}
  %where we sum over all the cost values from each location including the candidate location.
  %The major advantage of our model is that it can incorporate useful context information and learn the generation probabilities from data.
  Note that we do not set $g(l_s \rightarrow l_i)$ using simple distance functions, since we would like to learn more useful information from the observed trajectories. Typically, a user would select a route based on many factors. Our computation form for $g(\cdot)$ naturally fits the defined goal of our task in Eq.~\eqref{eq-prrtask}.
  To learn the neural network component, given  $g(\cdot)$ in Eq.~\eqref{eq-gdef}, we set a loss for all the observed trajectories over all users
  \begin{equation}\label{loss1}\small
  Loss_1 = \sum_{u \in \mathcal{U}} \sum_{p \in \mathcal{P}^u} g(p).
  \end{equation}
  %where $g(\cdot)$ is defined as in Eq.~\eqref{eq-gdef}.
  
  \subsection{Modeling the Estimated Cost with Value Networks}
  
  Besides the observable cost, we need to learn
   the estimated cost from a  candidate location to the destination.
  Specially, we introduce a value network to implement  $h(\cdot)$. This part is more difficult to model since no explicit trajectory information is observed.
  %The key problem is to estimate the cost between a candidate and the destination on a road network.
  %We take as input the road network, the current moving state of a user, a candidate location and the destination, and generate an estimated cost value.
  In order to better utilize the road network information for estimation, we build the value network on top of an improved graph attention network with useful context information.
  %Furthermore, since the optimal path from a candidate location to the destination involves a multi-step decision process, we further propose to use reinforcement learning for solving the model.
  
  \subsubsection{Improved  Graph Attention Networks for Road Networks}
  We consider using graph neural networks for learning effective structural representations for summarizing graph nodes. Generally, the update of graph neural networks~\cite{velickovic2017graph} can be given as
  \begin{equation}\small
  \bm{N}^{(z+1)}=\text{GNN}\left(\bm{N}^{(z)}\right)
  \end{equation}
  where $\bm{N}^{(z)} \in \mathbb{R}^{K_G \times |\mathcal{L}|}$ denotes the matrix consisting of node representations at the $z$-th iteration, and the $l_k$-th column $\bm{n}_{l_k} \in \mathbb{R}^{K_G}$ corresponds to
  the representation of node $l_k$, \ie location $l_k \in \mathcal{L}$.
  For initialization, we set $\bm{n}_{l_k}^{(0)}=\bm{v}_{l_k}$ with the learned location embeddings in Section 5.2.1.
  Here, we adopt the recently proposed Graph ATTention network (GAT)~\cite{velickovic2017graph} for a good balance between capacity and efficiency.
  In GAT, the key part is to compute the attention importance of a node on another. Original attention scores are computed using the node representations alone, which cannot well adapt to our task. %We make two meaningful extensions tailored for our task.
  
  \paratitle{Context-aware Graph Attention.} Intuitively, a node has a larger impact on a nearby node than a faraway node.
  For modeling this factor, we first discretize the distance $o_{l_i,l_j}$ between nodes $l_i$ and $l_j$ into consecutive value bins. Then we set a unique embedding $\bm{v}_o\in \mathbb{R}^{K_O}$ for each discretized distance value $o$. Besides, it is likely the previous part of a trajectory will affect the subsequent part. Recall that we use RNN to encode observable sub-trajectory $p: l_s \rightarrow l_i$ into an embedding vector $\bm{h}^{(p)}$, modeling the moving state of a user.
  The attention scores should take current moving state of a user into consideration.
  With the two factors considered, we compute the attention weights between two locations $l_{j}$ and $l_{j'}$ as
  \begin{equation}\small
  \alpha_{l_j,l_{j'}}=\frac{\exp\left(\bm{w}_2^{\top}\cdot\left(\bm{W}_3\bm{n}_{l_j}+\bm{W}_4\bm{n}_{l_{j'}}+\bm{W}_5\bm{h}^{(p)}+\bm{W}_6\bm{v}_{o_{l_j,l_{j'}}}\right)\right)}{\sum_{k \in \mathcal{L}_{l_j}}\exp\left(\bm{w}_2^{\top}\cdot\left(\bm{W}_3\bm{n}_{l_j}+\bm{W}_4\bm{n}_{l_k}+\bm{W}_5\bm{h}^{(p)}+\bm{W}_6\bm{v}_{o_{l_j,l_k}}\right)\right)},
  \end{equation}
  where $\bm{W}_{(\cdot)}$ and $\bm{w}_2$ are learnable parameters, and $\bm{v}_{o_{l_j,l_{j'}}}$ is the embedding vector for the discretized distance value.
  Since our attention mechanism involves more kinds of information, we  use the multi-head attention  for stabilizing the learning process. We combine the results of $A$ attention heads as
  \begin{equation}\label{eq-nodeembeddings}\small
  \bm{n}^{(z+1)}_{l_i} = \Big\Vert_{a=1}^A \mathrm{relu}\left(\sum_{l_j \in \mathcal{L}_{l_i}}\alpha_{l_i,l_j}^{(a)} \bm{W}^{(a)}\bm{n}_{l_j}^{(z)}\right),
  \end{equation}
  where $\alpha_{i,j}^{(a)}$ are the normalized attention scores computed by the $a$-th attention head, ``$\Vert$" denotes the concatenation operation and $\bm{W}^{(a)}$ is weight matrix of the corresponding input linear transformation.
  
  \ignore{
  \begin{equation}
  \alpha_{ij}=\frac{exp(\bm{a}^T[\bm{w}_h\bm{h}_i+\bm{w}_h\bm{h}_j+\bm{w}_c\bm{c}+\bm{w}_d\bm{d}_{ij}])}{\sum_{k \in \mathcal{N}_i}
  \bm{a}^T[\bm{w}_h\bm{h}_i+\bm{w}_h\bm{h}_k+\bm{w}_c\bm{c}+\bm{w}_d\bm{d}_{ik}]}
  \end{equation}
  where $\bm{w}_h$, $\bm{w}_c$ and $\bm{w}_d$ are parameters of linear transformation. These linear transformation are used to transform the input features, context information and spatial distance information into higher-level features. The spatial distance vector $\bm{d}_{ij}$ is an embedding vector gathered from distance embedding matrix according to the discretized value of spatial distance $d_{ij}$, and $c$ is the context vector that is the hidden state of GRU in satisfaction network that fuse attention context from user's history. To stabilize the learning process that involves so much information, inspired by the multi-head attention \cite{velickovic2017graph}, we extend our spatial-contextual attention mechanism to a multi-head spatial-contextual attention mechanism:
  }

  \paratitle{Predicting the Estimated Cost with MLP.} After obtaining the node representations, we are ready to define the cost function for estimating future cost.
  We use a  Multi-Layer Perceptron component to infer the cost from the candidate location $l_i$ to the destination $l_d$. Formally, we have
  \begin{equation}\label{eq-hcost}
  h\left(l_i \rightarrow l_d \right)= \text{MLP}\left(\bm{h}^{(p)}, \bm{n}_{l_i}, \bm{n}_{l_d}, \bm{v}_{o_{l_i,l_d}}\right),
  \end{equation}
  where the MLP component takes as input the moving state $\bm{h}^{(p)}$, the node representations $\bm{n}_{l_i}$ and $\bm{n}_{l_d}$ for locations of $l_i$ and $l_d$, and the embedding $\bm{v}_{o_{l_i,l_d}}$ for their spatial distance.

  \subsubsection{Temporal Difference Learning for the Estimated Cost}
  The  computation of  $h(l_i \rightarrow l_d )$ relies on the optimal sub-route from $l_i$ to $l_{d}$, which is a multi-step decision process and difficult to be directly optimized. Inspired by recent progress in reinforcement learning, we follow the framework of  Markov Decision Process~(MDP)~\cite{sutton2018reinforcement} for solving our problem. In an MDP, an agent behaves in an environment according to a policy that specifies how the agent selects actions at each state of the MDP.
   In our task, a state consists of the information of the query $q$ and the current sub-sequence $l_s \rightarrow l_{i-1}$;  an action is to select a location $l_i$ to extend in the route. The standard MDP aims to maximize the future reward, while
   our task  aims  to  minimize the future cost.
    Mathematically, maximizing the future reward is equal to minimizing the future cost for the PRR task.
  To be consistent with our task setting, in what follows, we model the \emph{cost} instead of \emph{reward}.
   %Different from original MDP, we focus on the learning of \emph{cost} but not \emph{reward}, and would like to minimize the sum of future cost.
   %Mathematically, maximizing the future reward is to equal to minimizing the future cost. Hence, in what follows, we model the \emph{cost} instead of \emph{reward}.
   % But it , which is   mathematically equal to maximize the sum of future reward.
  % Hence, we first define the immediate \emph{cost} at a step.
   When a location $l_i$ is selected, an immediate cost $c_i$ will be yielded according to
  \begin{equation}\small
  %\begin{aligned}
  c_i = - \log \text{Pr}\left(l_i |  l_s \rightarrow l_{i-1}, q, u\right),
  %\end{aligned}
  \end{equation}
  where $\text{Pr}(l_i | l_1 \rightarrow l_{i-1}, q, u)$ is the probability computed in Eq.~\eqref{eq-prli}.
  
  Since our purpose is to estimate the future cost, we do not need to explicitly learn the policy here. Hence, we adopt a popular value-based learning method for optimizing the value networks, \ie \emph{Temporal Difference (TD)}. TD~\cite{sutton1988learning} is an approach to learning how to predict a quantity that depends on future values of a given signal. In our model, the estimated cost $h\left(l_i \rightarrow l_d \right)$ and the future costs in $l_i \rightarrow l_d$ have a relationship as
  \begin{equation}\label{eq-hdef}\small
    h(l_i\rightarrow l_d) = \sum_{j=i+1}^{T}\gamma^{j-i-1}c_j,
  \end{equation}
  where $\gamma$ is discount rate to discounting future cost to the current and $T$ is the  timestamp arriving at $l_d$. If we look forward $n$ steps from the current location $l_i$, the prediction error of the value network is
  %\begin{equation}\small
  %\begin{aligned}
  %y_{l_i} &=& \gamma^n h(l_{i+n} \rightarrow l_d) + \sum_{j=i+1}^{i+n}\gamma^{j-i-1}c_{l_j},\\%\label{eq-TD-1}
  %\delta_{l_i} &=& \sum_{i = 1}^{T-1}\left\| h(l_i\rightarrow l_d) - y_{l_i} \right\|^2,%~\label{eq-TD-2}
  %\end{aligned}
  %\end{equation}
  \begin{eqnarray}\small
  y_{l_i} &=& \gamma^n h(l_{i+n} \rightarrow l_d) + \sum_{j=i+1}^{i+n}\gamma^{j-i-1}c_{l_j},\\\label{eq-TD-1}
  \delta_{l_i} &=& \sum_{i = 1}^{T-1}\left\| h(l_i\rightarrow l_d) - y_{l_i} \right\|^2,~\label{eq-TD-2}
  \end{eqnarray}
  where the term $y_{l_i}$ is the estimated cost using the temporal difference approach.
  %$[h(l_i\rightarrow l_d) - \gamma^n h(l_{i+n} \rightarrow l_d)]$ is the estimated cost of $l_i\rightarrow l_{i+n}$ using the temporal difference approach.
  To learn the value network component, we set a loss for all the observed trajectories over all users as
  \begin{equation}\label{loss2}\small
  Loss_2 = \sum_{u \in \mathcal{U}} \sum_{p \in \mathcal{P}^u} \sum_{l_i \in p} \delta_{l_i}.
  \end{equation}
  
  %Then we the backpropagation optimize the parameters in the value networks. Here, we follow the standard learning approach for TD with function approximation~\cite{}.

  \subsection{Model Analysis and Learning}
  Integrating the two components in Section 5.2 and 5.3, we obtain the complete NASR model for the PRR task. NASR follows the similar search procedure of $A^{*}$ algorithm but uses the learned cost for node evaluation.%~\footnote{See  Algorithm 2 in the Appendix for the NASR search procedure.}.
  Specially, it has fulfilled the cost functions of $A^{*}$ algorithms with neural networks, namely $g(\cdot)$ and $h(\cdot)$.
  Given a candidate location,
  the first component  utilizes RNNs to characterize the currently generated sub-trajectory for learning \emph{observable cost}, while the second component  incorporates a value network to predict the \emph{estimated cost} to arrive at the destination.
  Finally, the two cost values are summed as the final evaluation cost of a candidate location.
  
  %we  integrate the two components in a joint model for deriving the final evaluation cost.

  Compared with traditional heuristic search algorithms, NASR has the following merits.
  First, it does not require to manually set functions with heuristics, but automatically learns the functions from data.
  Second, it can utilize various kinds of context information and capture more complicated personalized trajectory characteristics.
  Third, it is able to coordinate and integrate the two components by sharing useful information or parameters in a principled way.
  Note that traditional search algorithms neglect the importance of $g(\cdot)$, which  computes the cost of observed sub-trajectories. In our model, the implementation of  $g(\cdot)$ not only learns the cost but also a vectorized user state representation, \ie the moving state of a user.
  This state vector is subsequently used for the learning of $h(\cdot)$ function by providing useful information from current sub-trajectory. Besides, as we discussed in Section 3, not all the observable cost can be directly computed, usually requiring estimation or approximation. Neural networks are helpful to improve the computation of $g(\cdot)$ by producing more robust results.
  
  To learn the model parameters, we first pre-train the RNN component. Then, we jointly learn the two components using alternative optimization by iterating over the trajectories in training set. After model learning, we follow the search procedure of $A^{*}$ algorithm for the PRR task with the evaluation cost computed by our model.
  \vspace{-0.2cm}

  \section{EXPERIMENTS}

  %In this section, we first set up the experiments, and then present the performance comparison and analysis.
  
  \subsection{Experimental Setup}

  %\begin{table}[t]\small
  %\caption{Statistics of the three datasets after preprocessing.
  %	}\label{tab:statistic}
  %	\centering
  %	\begin{small}
  %    %\setlength{\tabcolsep}{4mm}{
  %	\begin{tabular}{|c||c|c|c|}
  %		\hline
  %Statistics &  Beijing taxi & Porto taxi & Beijing bicycle \\
  %		\hline
  %		\hline
  %		%Duration & 1 month & 1 year & 1 month \\
  %		%\hline
  %        \#users & 18,298 & 442 & 196,591 \\
  %        \hline
  %		\#trajectories & 302,654 & 284,100 & 484,421 \\
  %		\hline
  %		\#records & 16,040,662 &  8,523,000 & 6,442,890 \\
  %		\hline
  %		\#locations & 15,208 & 8,224 & 15,500 \\
  %		\hline
  %		\#road segments & 20,198 & 9,457 & 22,010 \\
  %		\hline
  %	\end{tabular}
  %    %}
  %    \label{table1}
  %	\end{small}
  %	\vspace{-0.3cm}
  %\end{table}
  
  \begin{table*}[ttp]\small
  \caption{Performance comparison using four metrics on three datasets. All the results are better with larger values except the EDT measure. With paired $t$-test, the improvement of the NASR over all the baselines is significant at the level of 0.01.}
  \label{tab:summary}
  \centering
  %\begin{scriptsize}
   % \setlength{\tabcolsep}{3.0mm}{
    \begin{tabular}{|c||c||c|c|c|c|c|c||c|c|c|c|c|c|c|c|}
      \hline
          \multirow{2}{*}{Datasets}
      &\multirow{1}{*}{Metric}
      &\multicolumn{6}{c||}{\bf Precision}
      &\multicolumn{6}{c|}{\bf Recall}
      \\
      \cline{2-14}
      &Length&RICK&MPR&CTRR&STRNN&DeepMove&NASR&RICK&MPR&CTRR&STRNN&DeepMove&NASR
      \\
      \hline%\hline
  
          \multirow{2}{*}{Beijing}		
          &Short&0.712&0.347&0.558&0.491&0.742&\textbf{0.821}&0.723&0.372&0.164&0.384 &0.756&\textbf{0.848} \\
      \multirow{2}{*}{Taxi}
      &Medium &0.638&0.253&0.276&0.446&0.642 &{\bf 0.757} &0.651&0.261&0.067&0.350  &0.654&{\bf 0.773} \\
          &Long&0.586&0.169&0.194&0.359&0.562&{\bf 0.684}&0.589&0.173&0.045&0.214 &0.575&{\bf 0.709} \\
      \hline
          \multirow{2}{*}{Porto}		&Short&0.697&0.359&0.701&0.442&0.721&\textbf{0.804}&0.705&0.381&0.358&0.372  &0.726&{\bf 0.832}  \\
      \multirow{2}{*}{Taxi}
          &Medium&0.622&0.271&0.416&0.403&0.619&{\bf 0.729}&0.634&0.293&0.106&0.326  &0.628&{\bf 0.754}   \\
          &Long&0.565&0.184&0.305&0.340&0.547&{\bf 0.657}&0.578&0.198&0.036&0.218 &0.568&{\bf 0.671} \\
      \hline
          \multirow{2}{*}{Beijing}		&Short&0.652&0.303&0.587&0.559&0.673&\textbf{0.788}&0.670&0.313&0.272&0.330  &0.685&{\bf 0.802}  \\
          \multirow{2}{*}{Bicycle}
              &Medium&0.568&0.217&0.603&0.461&0.582&{\bf 0.715}&0.574&0.226&0.142&0.304  &0.589&{\bf 0.724}   \\
              &Long&0.503&0.129&0.613&0.297&0.487&{\bf 0.641}&0.519&0.139&0.045&0.206 &0.492&{\bf 0.663}  \\
      \hline
      \hline
          \multirow{2}{*}{Datasets}
      &\multirow{1}{*}{Metric}
      &\multicolumn{6}{c||}{\bf F1-score}
      &\multicolumn{6}{c|}{\bf EDT}
      \\
      \cline{2-14}
      &Length&RICK&MPR&CTRR&STRNN&DeepMove&NASR&RICK&MPR&CTRR&STRNN&DeepMove&NASR
      \\
      \hline%\hline
  
          \multirow{2}{*}{Beijing}		&Short&0.717&0.359&0.253&0.431&0.749&\textbf{0.834}&4.594&8.287&9.082&7.551 &4.362&{\bf 3.376}  \\
          \multirow{2}{*}{Taxi}
              &Medium&0.644&0.257&0.108&0.392&0.648&{\bf 0.765}&8.273&16.321&23.110&14.725  &8.730&{\bf 5.728}  \\
              &Long&0.587&0.171&0.073&0.268&0.568&{\bf 0.703}&11.283&25.873 &27.493&22.705 &12.059&{\bf 8.314} \\		\hline
          \multirow{2}{*}{Porto}		&Short&0.701&0.370&0.474&0.404&0.723&\textbf{0.818}&4.801&8.104&6.935&8.790  &4.496&{\bf 3.563} \\
          \multirow{2}{*}{Taxi}	
              &Medium&0.628&0.282&0.169&0.360&0.623&{\bf 0.741}&8.619&15.032&18.294&13.368  &8.930&{\bf 5.949} \\
              &Long&0.571&0.191&0.065&0.266&0.557&{\bf 0.687}&11.379&21.349&31.745&19.603 &12.297&{\bf 8.572} \\
      \hline
          \multirow{2}{*}{Beijing}		&Short&0.661&0.308&0.372&0.414&0.679&\textbf{0.795}&5.183&8.924&7.784&7.092  &4.629&{\bf 3.719}  \\
          \multirow{2}{*}{Bicycle}
              &Medium&0.571&0.221&0.229&0.367&0.585&{\bf 0.720}&8.972&17.497&20.966&14.503  &9.039&{\bf 6.253} \\
              &Long&0.511&0.134&0.084&0.243&0.489&{\bf 0.671}&11.891&22.028&57.997&21.324 &12.692&{\bf 8.794} \\
      \hline
    \end{tabular}
    %\end{scriptsize}
    \label{table2}
  \end{table*}

  \paratitle{Construction of the Datasets.}
  To measure the performance of our proposed model, we use three real-world trajectory datasets. The \emph{Beijing taxi} trajectory data  is sampled every minute, while the \emph{Beijing bicycle} dataset  is sampled every 10 seconds. The \emph{Porto taxi} dataset  is originally released for a Kaggle trajectory prediction competition  with a sampling period of 15 seconds. For the three datasets, we collect corresponding road network information from open street map.%~\footnote{\url{https://www.openstreetmap.org/}}.
  We further perform map matching~\cite{yang2018fast} by aligning GPS points with locations in the road network. In this way, we transform the trajectory data into timestamped location sequences.
  With the boundary indicators provided by the three datasets, we split the location sequence into multiple trajectories. %Table~\ref{tab:statistic} lists statistics of the three datasets after preprocessing. \emph{We will release the dataset and our source codes after the anonymous review.}

  \paratitle{Evaluation Metrics.}
  For the PRR task, we adopt a variety of evaluation metrics widely used in previous works~\cite{lim2015personalized, kurashima2010travel, cui2018personalized}.
  Given an actual route $p$, we predict a possible route $p'$ with the same source and destination. Following~\cite{lim2015personalized, cui2018personalized}, we use \emph{Precision}, \emph{Recall} and \emph{F1-score} as evaluation metrics: $Precision=\frac{|p \cap p'|}{|p'|}$,  $Recall=\frac{|p \cap p'|}{|p|}$ and $F1=\frac{2 * P* R}{P+R}$. \emph{Precision} and \emph{Recall} compute the ratios of overlapping locations \emph{w.r.t.} the actual and predicted routes respectively. Besides, we use the \emph{Edit distance} as a fourth measure~\cite{kurashima2010travel}, which is
  the minimum number of edit operations required to transform the predicted route into the actual route. Note  the source and destination locations are excluded in computing evaluation metrics.

  \ignore{First, we use \emph{Precision} and \emph{Recall} to measure the proportions of locations in the recommended route that are also contained in (1) the corresponding actual trajectory and (2)
  $R_u$.
  \item Recall~\cite{lim2015personalized, cui2018personalized} is the proportion of locations in a user's real-life trajectory $R_u$ that are also contained in the inferred most satisfied route $\hat{R}_u$.
  \item F1-score~\cite{lim2015personalized} is the harmonic mean of both the recall and precision of a inferred most satisfied route $\hat{R}_u$.
  \item Edit distance~\cite{kurashima2010travel} is the minimum number of edit operations required to transform the inferred most satisfied route $\hat{R}_u$ to its corresponding real route $R_u$.
  \end{itemize}
  }
  
  \paratitle{Task Setting.} For each user, we divide her/his trajectories into three parts with a  ratio of 7 : 1 : 2, namely training set, validation set and test set.
  We train the model with training set, and optimize the model with  validation set. Instead of reporting the overall performance on all test trajectories, we generate three types of queries \emph{w.r.t.} the number of locations in the trajectories, namely
  \emph{short} (10 to 20 locations), \emph{medium} (20 to 30 locations) and \emph{long} (more than 30 locations). In test set, given a trajectory, the first and last locations are treated as the source and destination respectively, and the rest locations are hidden.
  Each comparison method is required to recover the missing route between the source and destination.

  \ignore{
  Then, stop points(stayed on a road link for more than one timestamp) will be removed. After processing, every dataset will be divided into three parts with the splitting ratio of 7 : 1 : 2, namely training set, validation set and test set. For every user, we collect part of his trajectories in a period time as the input of historical attention module, which is shared in training, validating and testing stages. In test phrase, we randomly generate three kinds of query: short, medium and long according to the number of locations between source and destination in actual trajectory. In our setting, short query means the range of trajectory's raw length is 5 to 10, medium query means a range of 10 to 20, and long query means that the raw length is longer than 20.
  }
  
  \paratitle{Methods to Compare.} We consider the following  comparisons:
  
   \textbullet \textit{RICK}~\cite{Wei2012Constructing}: It builds a routable graph from uncertain trajectories, and then answers a users online query (a sequence of point locations) by searching top-$k$ routes on the graph.
  
  \textbullet \textit{MPR}~\cite{Chen2011Discovering}: It discovers the most popular route from a transfer network based on the popularity indicators in a breadth-first manner.
  
  \textbullet \textit{CTRR}~\cite{cui2018personalized}: It proposes collaborative travel route recommendation by  considering user's personal travel preference.
  
  %\textbullet \textit{RankMarkov}~\cite{chen2016learning}: It proposes an approach to recommending trajectories by jointly optimizing point preferences and routes.
  \textbullet \textit{STRNN}~\cite{Liu2016Predicting}: Based on RNNs, it models local temporal and spatial contexts in each layer with transition matrices for different time intervals and geographical distances.
  
  \textbullet \textit{DeepMove}~\cite{feng2018deepmove}:  It is a multi-modal embedding RNN that can capture the complicated sequential transitions by jointly embedding the multiple factors that govern the human mobility.

  Among these baselines, RICK and MPR are heuristic search based methods, CTRR is a  machine learning method, and STRNN and DeepMove are deep learning methods. The parameters in all the models have been optimized using the validation set. %We leave the parameter configuration and tuning  in the appendix.
  %\vspace{-0.5cm}
  
  \subsection{Results and Analysis}
  
  We present the results  of all the comparison methods in Table \ref{table2}.
  First, heuristic search methods, \ie RICK and MPR,  perform very well, especially the RICK method. RICK fully characterizes the road network information and adopts the informed $A^{*}$ algorithm. As a comparison, MPR mainly considers the modeling of transfer network  and uses a relatively simple BFS search procedure.
  Second, the matrix factorization based method CTRR does not perform better than RICK and MPR.
  A possible reason is that CTRR can not well utilize the road network information.
  Besides, it has limited capacities in learning complicated trajectory characteristics. In our experiments, CTRR tends to generate short route recommendations,  giving very bad recall results for medium and long queries.  Third, deep learning method DeepMove performs very well among all the baselines, while STRNN gives a worse performance.
  Compared with STRNN, DeepMove considers more kinds of context information and designs more advanced sequential neural networks.
  Finally, the proposed model NASR is consistently better than all the baselines in all cases, yielding very good performance even on long queries.
  
  By summarizing these results, we can see heuristic search methods are competitive to solve the PRR task, especially when suitable heuristics are used and context information is utilized. Besides,  deep learning is also able to improve the performance by leveraging the powerful modeling capacity.
  Our proposed model NASR is able to combine both the benefits of heuristic search and neural networks, and hence it performs best among the comparison methods.
  
  %As we have mentioned, NASR is able to combine both the benefits of the $A^{*}$ search and neural networks.
  
  \subsection{Detailed Analysis on Our Model NASR}
  
  In this section, we perform a series of detailed analysis on NASR for further verifying its effectiveness. Due to space limit, we only report the results of F1 scores on the \emph{Beijing taxi} dataset. The rest results show the similar findings, and are omitted here.

  \begin{figure}[t]
    \centering
    \subfigure[{Examining the RNN component.}]{
      \label{fig:subfig:music} %% label for first subfigure
      \includegraphics[width=0.15\textwidth]{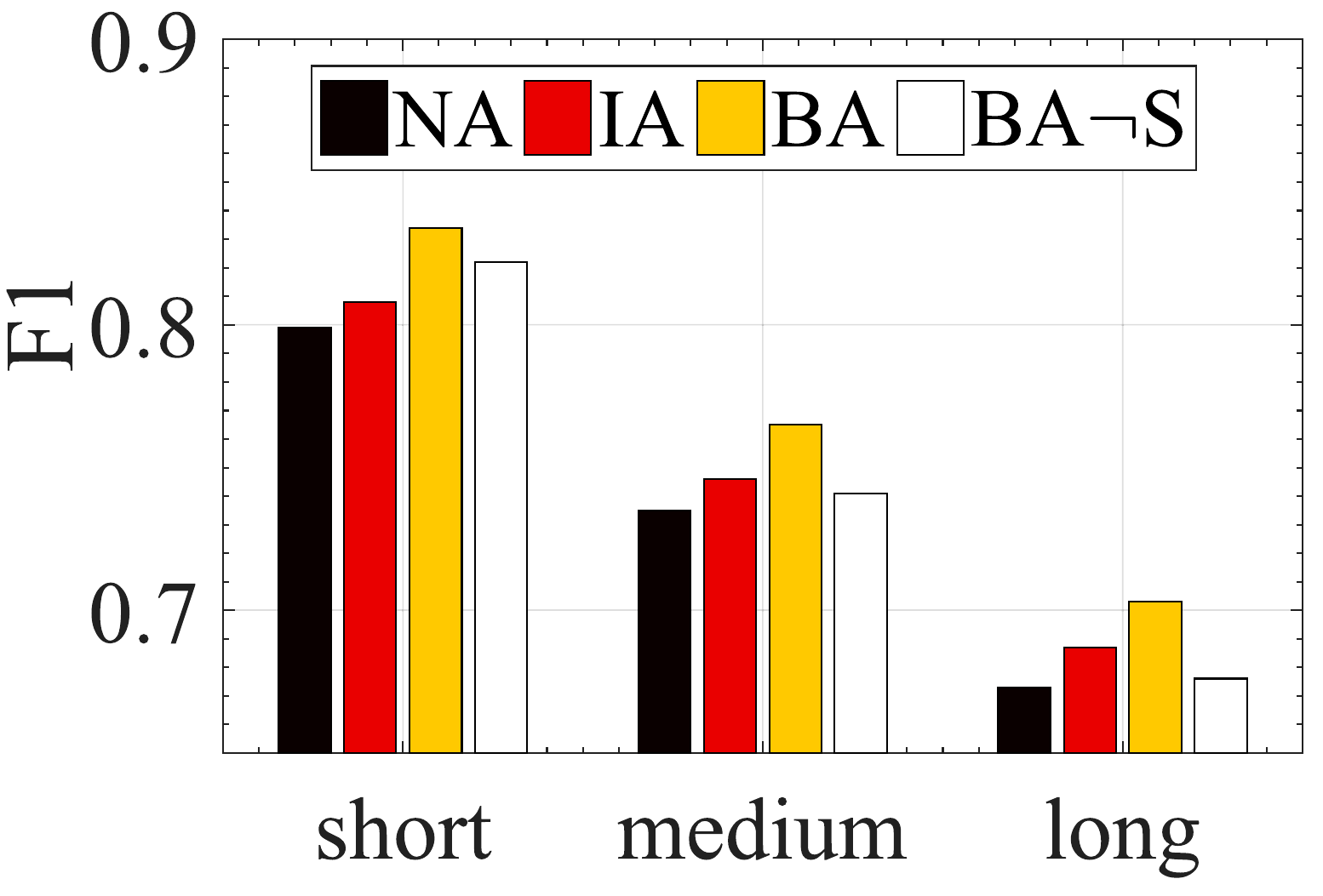}}
        \subfigure[{Examining the value network.}]{
      \label{fig:subfig:music} %% label for first subfigure
      \includegraphics[width=0.15\textwidth]{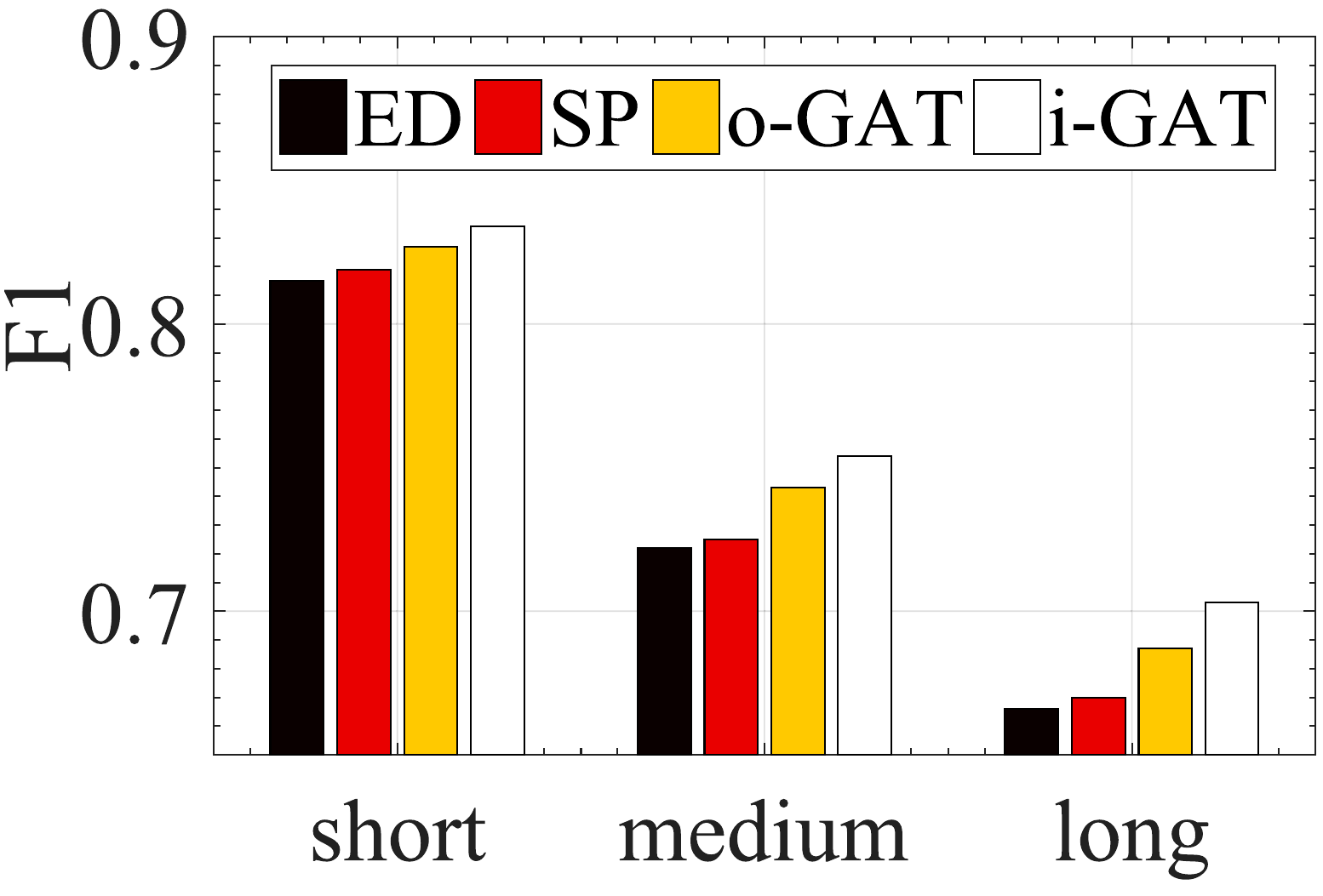}}
    \subfigure[{Examining the TD learning method.}]{
      \label{fig:subfig:book} %% label for second subfigure
      \includegraphics[width=0.15\textwidth]{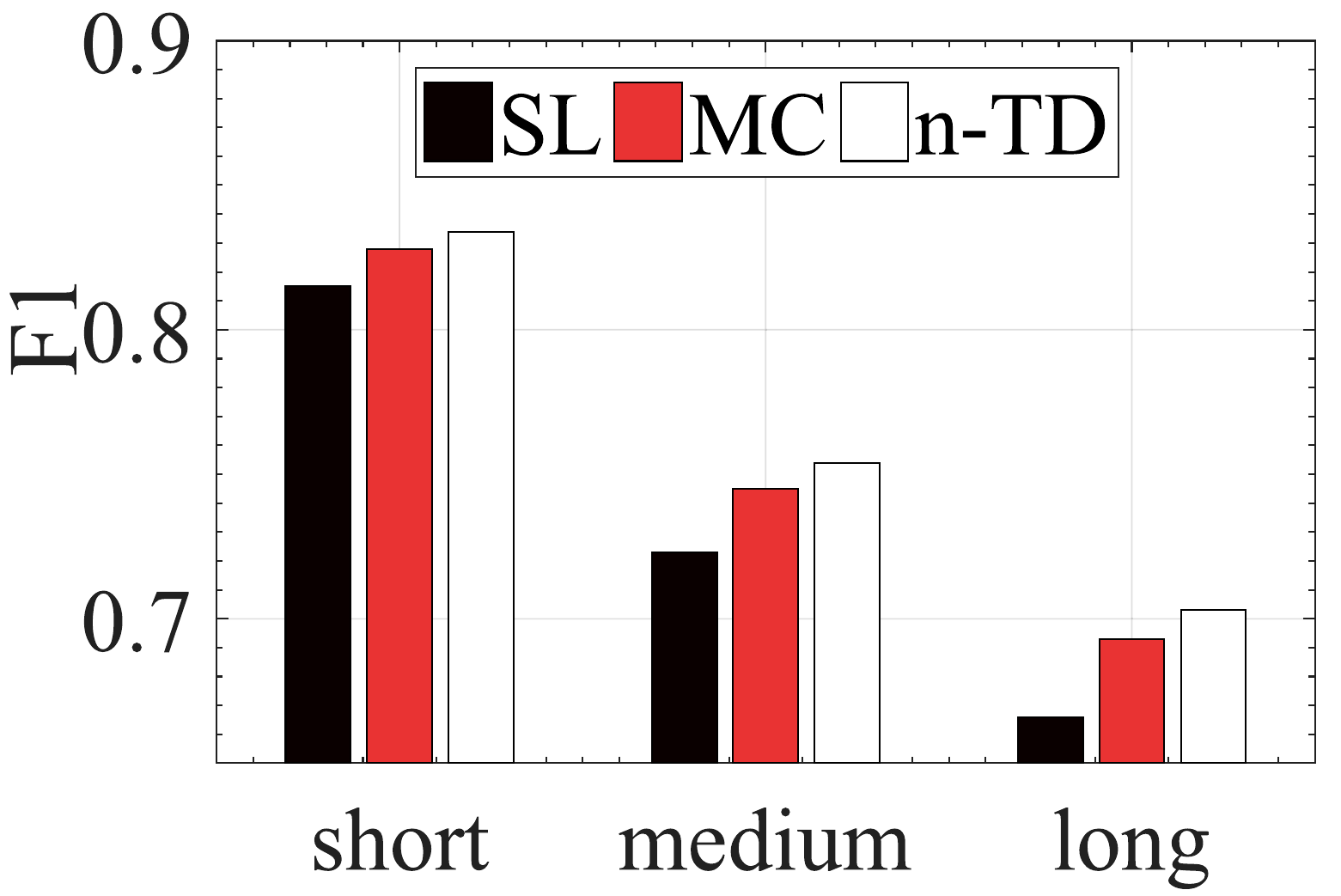}}
  %    \subfigure[{Examining the \# of training data.}]{
  %    \label{fig:subfig:book} %% label for second subfigure
  %    \includegraphics[width=0.223\textwidth]{figures/exp4.pdf}}
      \vspace{-0.4cm}
    \caption{Detailed analysis of our model on the dataset of Beijing taxi using F1 measure.}
    \label{fig-results} %% label for entire figure                         
  \end{figure}

  \paratitle{Effect of the RNN Component.}  We first examine the effect of the RNN component with different variants.
  We have incorporated two kinds of attentions, namely inter- and intra-trajectory attention in Section 5.2. Here, we consider three variants of the attention mechanism for implementing $g(\cdot)$: \emph{without attention (\underline{NA})}, \emph{using only intra-trajectory attention (\underline{IA})} and \emph{using both intra- and inter-trajectory attention (\underline{BA})}. %For a direct comparison, we do not use the value network in the three variants, and set $f=g$. In this case, the three variants degenerate into depth-first search algorithms.
  Recall our RNN component is also able to learn a vectorized representation for the moving state of users. We further prepare a variant for verifying the effect of the learned moving state in the value network, namely the model that does not provide the moving state to the $h(\cdot)$ function, denoted by (\underline{\emph{BA}$_{\neg S}$}).
  In Fig.~\ref{fig-results}(a), it can be seen that the performance rank is as follows:
  \emph{NA $<$ IA$<$ BA} and \emph{BA$_{\neg S}$ $<$ BA} . It shows that both inter- and intra-trajectory attention are important  to improve the performance of the PRR task.
  Especially,  the learned moving state from the RNN component is useful for the value network. When the moving state is incorporated, the performance of the joint model has been substantially improved.%  It indicates the importance of the learned moving state in our task.

  \paratitle{Effect of the Value Network}. Predicting the estimated cost (\ie $h(\cdot)$) of a candidate location is especially important for our task. We use a value network for implementing $h(\cdot)$, which replaces the traditional heuristics. We now examine the
  performance of different variants for the value network. In this part, we fix the RNN component as its optimal setting. Then we prepare four variants for the value network as comparisons, including (1) \underline{\emph{ED}} using Euclid distance as heuristics, (2) \underline{\emph{SP}} using the scalar product between the embeddings of the candidate  and destination locations, (3)
  \underline{\emph{o-GAT}} using the original implementation of graph attention networks, and (4) \underline{\emph{i-GAT}} using our improved GAT by incorporating context information.  Both variants (3) and (4) are trained using the same TD learning method.
  In Fig.~\ref{fig-results}(b), it can be observed that the performance rank is as follows:
  \emph{ED} $<$  \emph{SP} $<$ \emph{o-GAT} $<$ \emph{i-GAT}. We can see that the simplest spatial distance baseline \emph{ED} gives the worst performance, which indicates simple heuristics may not work well in our task. Graph attention networks are more effective to capture structural characteristics from graphs. When incorporating context information, our value network is able to outperform the variant using original implementation.

  \paratitle{Effect of Temporal Difference Learning Method}. To learn our model, an important technique we apply is the Temporal Difference (TD) method. For verifying the effectiveness of the $n$-step TD method, we consider four variants for comparison, including (1) \underline{\emph{SL}} which directly learns the actual distance between the candidate location and the destination in a supervised way, (2) \underline{\emph{MC}} which applies Monte Carlo method to generate sampled sequences and trains the model with the cost of these sampled sequences~\cite{sutton2018reinforcement}, (3)   \underline{\emph{$n$-TD}}  which uses a TD step number of $5$.
  From Fig.~\ref{fig-results}(c), we can see that the simplest supervised learning method performs worst.
  Since the prediction involves multi-step moving process, it is not easy to directly fit the distance using traditional supervised learning methods.
  Compared with all the methods, we can see that the $5$-TD learning method is the most effective in our task. In our experiments, we find that using a step number of 5
  produces the optimal performance.

  \begin{figure}[t]
   \includegraphics[width=0.8\columnwidth]{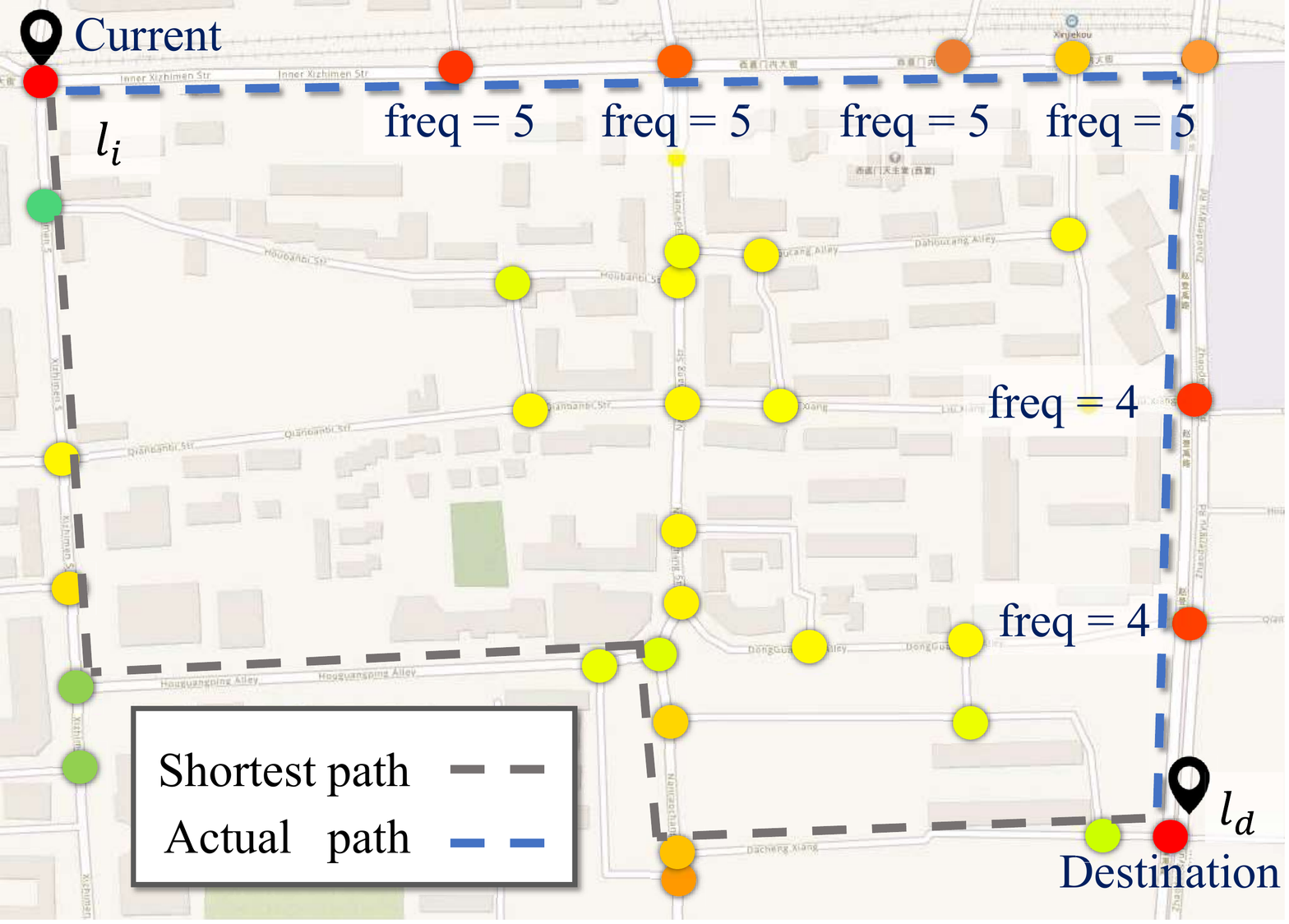}\vspace{-.3cm}
   \caption{Visualization of  the learned association scores using  improved graph attention networks. The colored circles denote locations in the road network. A darker color indicates a larger importance degree \emph{w.r.t.} current location $l_i$ and destination $l_d$. ``\emph{freq}" denotes the visit frequency by the user in historical trajectories.%We also report the actual visiting frequencies.
   }
   \label{fig-gat}\vspace{-.3cm}
  \end{figure}

  \subsection{Qualitative Analysis}
  Previously, we have shown the effectiveness of our model in the PRR task. In this part, we qualitatively analyze why NASR is able to yield a good performance.
  
  In NASR, the improved graph attention network is the core component for
  modeling road network information. It can generate informative node representations for encoding structural characteristics. To see this, we present an illustrative example in Fig.~\ref{fig-gat}. A user is currently located at $l_i$ and moving towards the destination $l_d$. For a candidate location $l_j$, we compute a simple scoring formula: $\bm{n}_{l_j}^{\top} \cdot \bm{n}_{l_i} + \bm{n}_{l_j}^{\top} \cdot \bm{n}_{l_d}$, where $\bm{n}_{(\cdot)}$s
  are the node representations learned in Eq.~\eqref{eq-nodeembeddings}.
  This formula measures the association degree of $l_j$ with both current location and destination.
  For comparison, we plot both the actual and shortest route.
  As we can see, the locations on the actual route has a larger association weight than those on the shortest route.
  By inspecting into the dataset, we find the shortest route contains several side road segments that are possibly in traffic congestion at the visit time. Another interesting observation is that the user indeed visits the locations in the actual route more times in historical trajectories.
  These observations indicate that our model is able to learn effective node representations for identifying more important locations to explore for the PRR task.

  %capture multiple kinds of useful information for deriving a more accurate association between locations for the PRR task.

   %For examining the effectiveness of the learned attention weights, we also plot the actual route and the shortest route. As we can see, the points on the actual route has a larger attention weight than those on the shortest route.
  
\begin{figure}[t]
   \includegraphics[width=0.97\columnwidth]{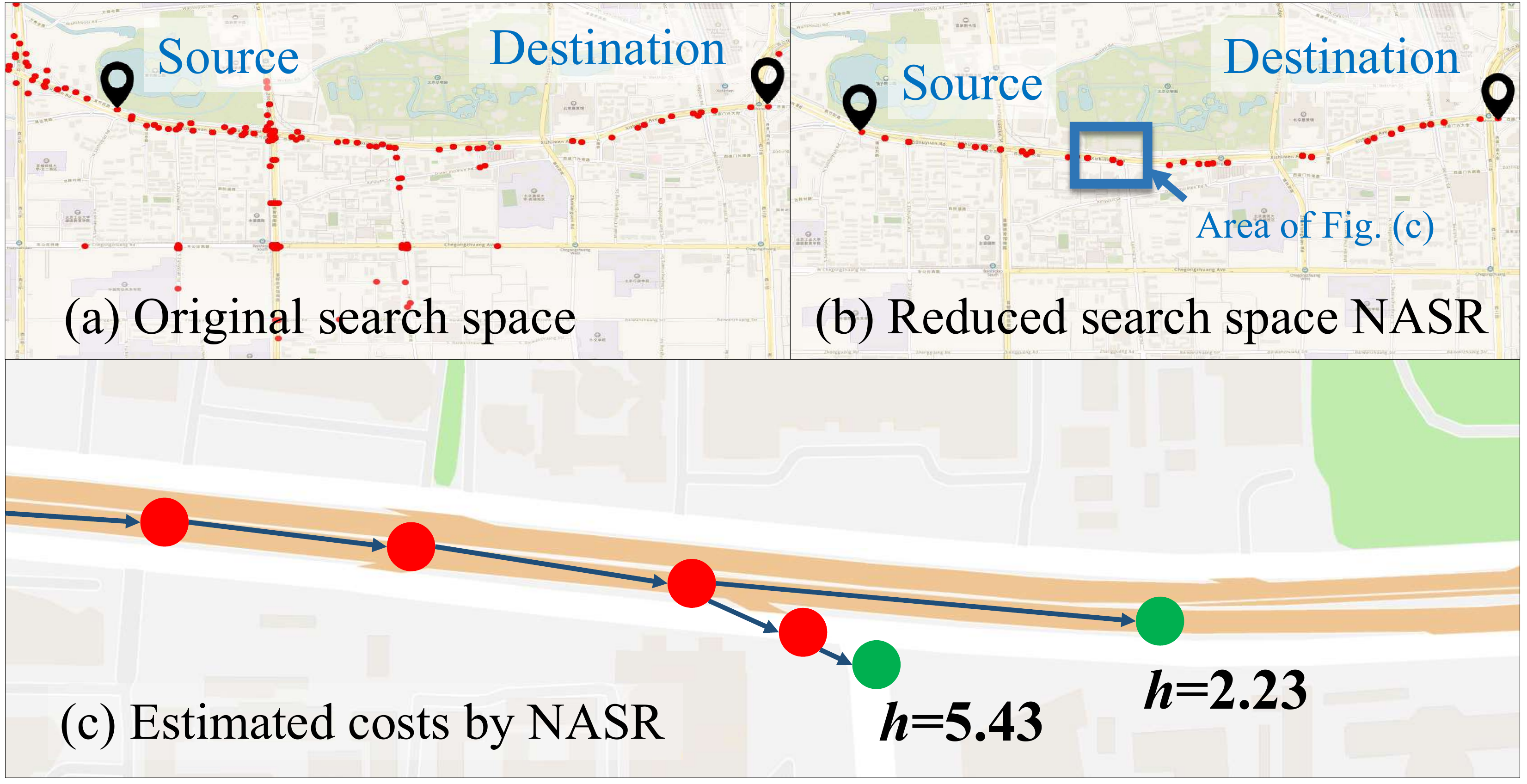}\vspace{-.3cm}
   \caption{Visualization of the search procedure with the estimated costs by the NASR model. In (c), red points have been already explored and green points are candidate locations to extend in  $A^{*}$ search algorithm.}
   \label{fig-search}\vspace{-.3cm}
  \end{figure}

  Next, we continue to study how the learned cost function helps the search procedure in NASR. Figure~\ref{fig-search} presents a sample trajectory  from a specific user.
  Given the source and destination, we need to predict the actual route.
  By comparing Fig.~\ref{fig-search}(a) (the original search space) and Fig.~\ref{fig-search}(b) (the reduced search space by NASR), it can be seen that our model is able to effectively reduce the search space. When zooming into a subsequence of this route, we  further compare the estimated cost values for two candidate locations (green points) in Fig~\ref{fig-search}(c).
   Although the second location has a longer distance with the explored locations, it is located on the main road that is likely to lead to a better traffic condition. Our model is able to predict a lower cost
  for the second location by effectively learning such trajectory characteristics from road network and historical data.

  \section{CONCLUSIONS}
  In this paper, we took the initiative to use neural networks to automatically learn the cost functions in $A^{*}$ for the PRR task. We first presented a simple $A^{*}$ solution for solving the PRR task, and formally defined the suitable form for the search cost. Then, we set up two components to learn the two costs respectively, \ie the RNN component for $g(\cdot)$ and the value network for $h(\cdot)$. The two components were integrated in a principled way for deriving a more accurate cost of a candidate location for search.
  We constructed extensive experiments for verifying the effectiveness and robustness of the proposed model.
  
  Since road network information is not always available, as future work, we will consider extending our model to solve the PRR task without road networks. 
  Currently, we  focus on the PRR task. 
  We will also study whether our solution can be generalized to solve other complex search  tasks. 
  %With road network information, our model 
  %As future work, we plan to investigate into the theoretical analysis for our proposed model. Since the cost functions are learned using neural networks, it is difficult to directly analyze some important properties of $A^{*}$, \eg optimality and admissibility. In addition, we currently select $A^{*}$ as the base search algorithm for the PRR task, and it will be meaningful to study whether the proposed model can be generalize to solve other tasks. 

%\acknowledgments{the National Natural Science Foundation of China (Grant No.61572059)}

\section*{Acknowledgments}
The work was partially supported by National Natural Science Foundation of China under the Grant Number 61572059, 71531001, 61872369 and  the Fundamental Research Funds for the Central Universities. Wang's work was partially supported by the National Key Research and Development Program of China under Grant No. 2016YFC1000307. Zhao's work was supported by the Research Funds of Renmin University of China under Grant No. 18XNLG22. Wu's work was supported by the Science and Technology Project of Beijing under Grant No.Z181100003518001, and the Open Foundation of TUCSU under Grant No.TUCSU-K-17002-01.

%\bibliographystyle{ACM-Reference-Format}
%\bibliography{reference}
%%% -*-BibTeX-*-
%%% Do NOT edit. File created by BibTeX with style
%%% ACM-Reference-Format-Journals [18-Jan-2012].

\end{document}